\documentclass{article}

\usepackage[preprint]{neurips_2019}

\usepackage[]{neurips_2019}

\usepackage[utf8]{inputenc} 
\usepackage[T1]{fontenc}    
\usepackage{hyperref}       
\usepackage{url}            
\usepackage{booktabs}       
\usepackage{amsfonts}       
\usepackage{nicefrac}       
\usepackage{microtype}      
\usepackage{graphicx}
\graphicspath{ {./Images/} }
\usepackage{subfig}

\title{Exploiting the potential of deep reinforcement learning for classification tasks in high-dimensional and unstructured data}

\author{%
  Johan S. Obando-Ceron\\
  Department of Automatics and Electronics\\
  Universidad Autónoma de Occidente\\
  \texttt{jsobando@uao.edu.co} \\
  \And
     Victor Romero Cano\\
 Department of Automatics and Electronics \\
  Universidad Autónoma de Occidente\\
  \texttt{varomero@uao.edu.co} \\
   \And
      Walter Mayor Toro\\
  Department of Automatics and Electronics\\
  Universidad Autónoma de Occidente\\
  \texttt{wmmayor@uao.edu.co} \\
}

\begin{document}
\maketitle
\begin{abstract}

This paper presents a framework for efficiently learning feature selection policies which use less features to reach a high classification precision on large unstructured data. It uses a Deep Convolutional Autoencoder (DCAE) for learning compact feature spaces, in combination with 
recently-proposed Reinforcement Learning (RL) algorithms as Double DQN and Retrace.


\end{abstract}

\section{Research problem and motivation}

RL has become a very powerful  technique  in  the  last  decade  thanks  to  multiple  technological
advances in computational  power  and  efficient  learning  algorithms  in  neural networks. However, RL for high-dimensional spaces  is  far  from being applied in real  world  applications  due  to  many
computational challenges [1]. 

As it is known, RL  allows
an   autonomous   agent   to   learn   through   experience   how   to
solve  a  problem  in  an  optimal  way,  with  minimal  information  about  its  environment. Some of the most popular modern RL algorithms,  use deep neural networks and back-propagation in order to maximize a reward function which indicates
how well the agent is performing in the environment [2]. However,
when these algorithms are applied to high-dimensional data such
as high-resolution images, they implicitly must learn to extract useful
features and their computational complexity increases. Therefore, in  this  paper we present a framework that harnesses the power of deep learning approaches for feature learning, in particular we discuss the effectiveness of DCAE in a pattern recognition task like image classification using RL. 

For classification problems, Deep Reinforcement Learning (DRL) has served in eliminating noisy data and learning better features, which made a great improvement in classification performance for structured datasets [3, 4, 5, 6]. However, there has been little research work on adapting these methods so they can be applied to unstructured data like images. Additionally, there is no substantial work focused on applying DCAE to the problem of image classification tasks using DRL to accelerate its learning process.


The considerations above motivate the formulation of this work’s research question: how to develop an efficient DRL framework that is able to deal with high-dimensional unstructured data and provides a sound method for optimal feature selection in image classification tasks?



\section{Technical contribution}
\label{gen_inst}

To address the problem of image classification, we use a DCAE to
extract features automatically. Subsequently, we formulate the classification problem
as a sequential decision-making process, which is solved by a deep Q-learning network (DQN) [7]. DQN is a RL technique used for learning a classification policy in which an agent has the ability to select the most helpful subset of features and filter the irrelevant or redundant ones. The goal of the agent is to obtain
as much cumulative rewards as possible during the process of
sequential decision-making, that is, to correctly recognise the
samples in a consistent manner.

In this formulation, each
sample corresponds to an episode, where an agent sequentially decides whether to acquire another feature and which feature to acquire, or if it is possible to already classify the sample. At each time step, the agent receives
an environment state which is represented by a training sample
and then performs a classification action under the guidance
of a policy. If the agent performs a correct classification action
it will be given a positive reward, otherwise, it will be given
a negative reward. For the actions requesting a new feature,
the agent receives a negative reward, equal to the feature cost. 

Hand-crafting features may neglect some potential helpful information leading to the classification model not having a stable behaviour. Therefore, using DCAE and DRL rather than classical classification algorithms poses great benefits. Our method uses less features to reach a relatively higher classification precision.  The DRL community has made several independent improvements to the DQN algorithm. However, it is unclear which and how these extensions can be fruitfully applied to the problem of classifying high-dimensional data. Therefore, in this paper, we propose to combine an extended DRL algorithm called Double DQN [8] together with retrace [9], and a DCAE. Fig. 1(a) shows our proposed model called Autoencoder Double Deep Q-Network (AE-DDQN).

\begin{figure}[ht]
  \subfloat[]{
	\begin{minipage}[c][0.42\width]{
	   0.5\textwidth}
	   \centering
	   \includegraphics[width=0.9\textwidth]{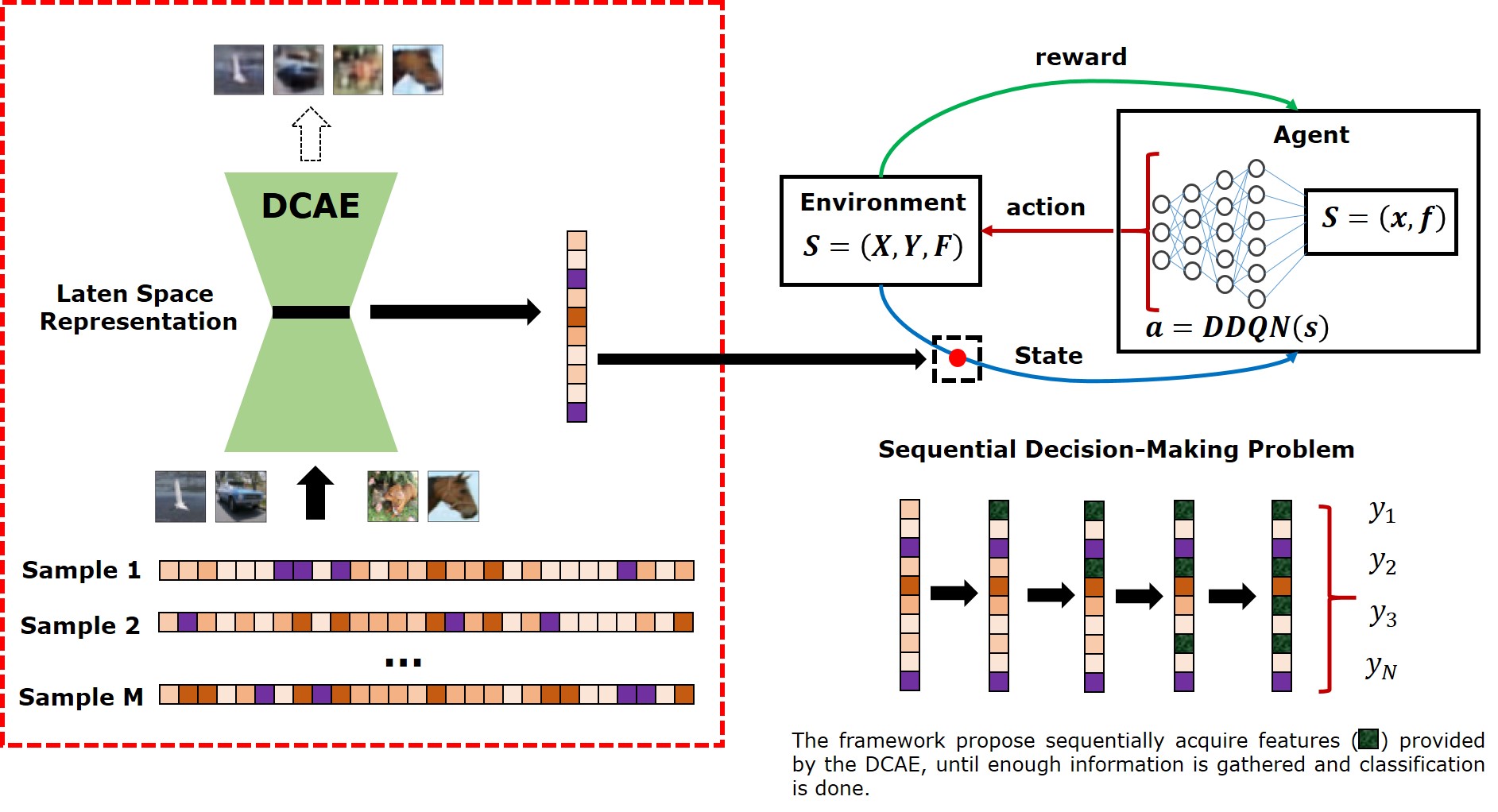}
	\end{minipage}}
 \hfill 	
  \subfloat[]{
	\begin{minipage}[c][0.42\width]{
	   0.5\textwidth}
	   \centering
	   \includegraphics[width=0.70\textwidth]{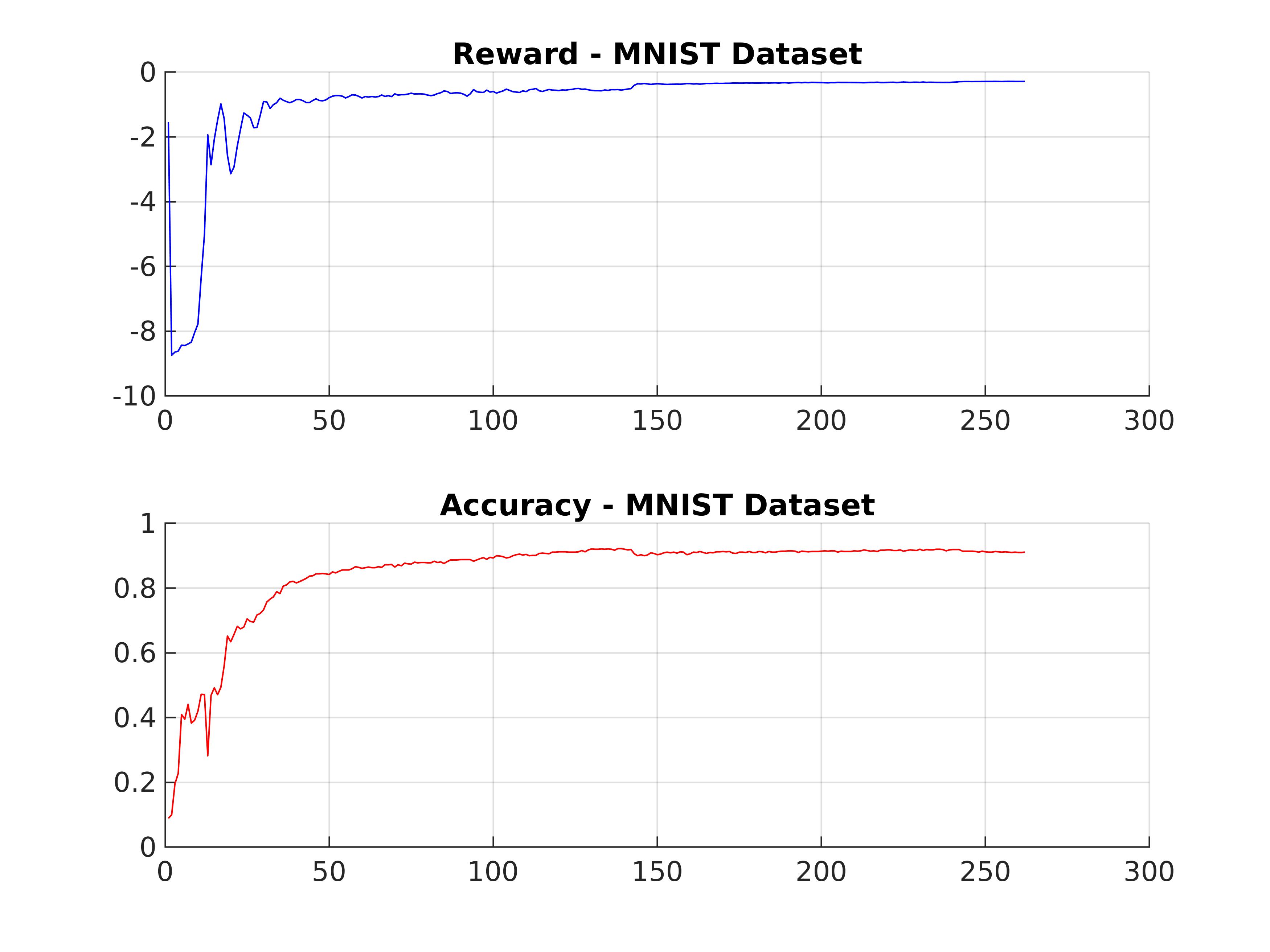}
	\end{minipage}}
\caption{(a) General framework proposed. (b) Accuracy and Average reward during training process.}
\end{figure}

AE-DDQN takes care of choosing an optimal subset of the features provided by the autoencoder to reach a relatively high precision. To evaluate it, we use the result of an SVM classifier as a baseline whose average precision is about 86.8\% for the MNIST [10] classification task. As it shown in Fig. 1(b), our classification method's accuracy is about
92.2\%, approximately six percent points higher than the baseline, and it uses less features than common classification algorithms. The average reward and the accuracy during the training process are shown in Fig. 1(b).

\begin{table}[!htb]
  \centering
\caption{Comparison of the accuracy for SVM and AE-DDQN}
\label{sample-table}
\begin{tabular}{p{0.6cm} p{0.8cm} c p{0.3cm} c c c}
\toprule
\multicolumn{1}{ c }{ \textbf{ \ Dataset}} & \textbf{Feats.} & \textbf{Feats. (DCAE)} & \textbf{Cls.}  & \textbf{Acc. (SVM)}  & \textbf{Acc. (Ours)} & \textbf{\# Feats used.} \\ \cline{1-7}

\multicolumn{1}{ c }{MNIST} & 784 & 128 & 10 & 86.8\% & 92.2\% &112   \\ 
\multicolumn{1}{ c}{CIFAR-10} & 1024 & 256& 10&  88\% & 81.3\%  & 198\\ 
\multicolumn{1}{ c }{Fashion-MNIST} & 784 & 128 & 10 & 88.3\% & 93.1\% & 98 \\  
\bottomrule
\end{tabular}
\end{table}

Fig. 1(b) shows how the accuracy increases quickly at the beginning and converges to a stable value slowly after reaching a high precision. High precision occurs once the agent finds a minimum subset of important features. By selecting valuable features we obtain an improved classifier which in turn feeds back a higher reward to the DDQN. The higher reward encourages the agent to select more advantageous features without forgetting their cost. It's very important to take into account that these datasets do not have any costs associated, hence we treat them with equal importance, assigning the same cost for all of the features.  Experiments on different datasets show that our proposed model outperforms vanilla classification algorithms like SVM as shown in Table 1. On a large dataset like CIFAR-10 [11], although the classification performance was not improved, the number of selected features was less. This can be considered as positive result in scenarios where low amounts of data is available.

\section*{References}
\small

[1] Daiki Kimura. 2018. DAQN: Deep Auto-encoder and Q-Network.
arXiv preprint
arXiv:1806.00630, 2018.

[2] Hessel, M., Modayil, J., van Hasselt, H., Schaul, T., Ostrovski, G., Dabney, W., Horgan, D., Piot,
B., Azar, M., and Silver, D. Rainbow: Combining Improvements in Deep Reinforcement
Learning. In
the AAAI Conference on Artificial Intelligence (AAAI), 2018.

[3] Dulac-Arnold, G.; Denoyer, L.; Preux, P.; and Gallinari, P. Datum-wise classification: a sequential approach to sparsity. In
Joint European Conference on Machine Learning and Knowledge
Discovery in Databases, 375–390, 2011.

[4] J. Janisch, T. Pevny, and V. Lisy, Classification with costly features
using deep reinforcement learning, arXiv preprint arXiv:1711.07364, 2017.

[5] Lin, E., Chen, Q., and Qi, X, Deep reinforcement learning for imbalanced classification, preprint https://arxiv.org/abs/1901.01379, 2019.

[6] Kachuee, M., Karkkainen, K., Goldstein, O., Zamanzadeh, D., and Sarrafzadeh, Majid., Cost-Sensitive Diagnosis and Learning Leveraging
Public Health Data, preprint https://arxiv.org/abs/1902.07102, 2019.

[7] V. Mnih, K. Kavukcuoglu, D. Silver, A. A. Rusu, J. Veness,
M.  G.  Bellemare,  A.  Graves,  M.  Riedmiller,  A.  K.  Fid-
jeland,  G.  Ostrovski,  S.  Petersen,  C.  Beattie,  A.  Sadik,
I. Antonoglou, H. King, D. Kumaran, D. Wierstra, S. Legg,
and D. Hassabis. Human-level control through deep rein-
forcement learning.
Nature, 2015.

[8] H. van Hasselt, A. Guez, and D. Silver. Deep reinforcementlearning with Double Q-learning.AAAI, 2016b.

[9] Rémi Munos, Tom Stepleton, Anna Harutyunyan, and Marc G Bellemare.  Safe and efficient Off-Policy reinforcement learning. InAdvances in Neural Information Processing Systems, 2016.

[10] Y. LeCun, L. Bottou, Y. Bengio, and P. Haffner, “Gradient-based learning
applied  to  document  recognition,”
Proceedings  of  the  IEEE
,  vol.  86,
no. 11, pp. 2278–2323, 1998.

[11] A.  Krizhevsky,  “Learning  Multiple  Layers  of  Features  from  Tiny  Im-
ages,” Ph.D. dissertation, University of Toronto, 2009.

\end{document}